\documentclass[final,3p,times,twocolumn]{elsarticle}
\usepackage{amssymb}
\usepackage{hyperref}
\journal{}

\begin{document}

\begin{frontmatter}



\title{COVIDCalc: A Novel Calculator to Measure Positive U.S. Socio-Economic Impact of a COVID-19 AI Pre-screening Solution}


\author[inst1]{Richard Swartzbaugh}

\affiliation[inst1]{organization={CuraeSoft Corporation},
}

\author[inst2]{Amil Khanzada}
\affiliation[inst2]{organization={University of California, Berkeley}}
\author[inst3]{Jennifer Ranjani J}
\affiliation[inst3]{organization={Virufy}}
\author[inst4]{Praveen Govindan}
\affiliation[inst4]{organization={Indian Institute of Technology, Madras}}

\author[inst5]{Mert Pilanci}
\affiliation[inst5]{organization={Stanford University}}

\author[inst6]{Ayomide Owoyemi}
\affiliation[inst6]{organization={ University of Illinois, Chicago}}

\author[inst7]{Les Atlas}
\affiliation[inst7]{organization={ University of Washington}}

\author[inst8]{Hugo Estrada}
\affiliation[inst8]{organization={ Iwe Consultores}}

\author[inst9]{Richard Nall}
\affiliation[inst9]{organization={ The Brand Garden}}

\author[inst10]{Michael Lotito}
\affiliation[inst10]{organization={ Littler Mendelson}}

\author[inst10]{Rich Falcone}

\begin{abstract}
The COVID-19 pandemic has been a scourge upon humanity, claiming the lives of more than 5.1 million people worldwide; the global economy contracted by 3.5\% in 2020. Several calculators have been developed to estimate the cost associated with sectors such as like health care, education. This paper presents a COVID-19 calculator, synthesizing existing published calculators and data points, to measure the positive U.S. socio-economic impact of an AI solution that can be used to pre-screen COVID-19 using the cough samples. This calculator provides its users an opportunity to include their own data. 
\end{abstract}


\end{frontmatter}


\section{Introduction}
\label{sec:sample1}
In recent history, the singular event that most profoundly impacted the entire globe has been the COVID-19 pandemic. To date, COVID-19 has claimed the lives of over 800,000 individuals in the United States and 5 million worldwide \cite{1}. The highly infectious nature of COVID-19 has filled up hospital beds in record numbers, surpassing hospital capacity and causing immense strain on healthcare systems worldwide \cite{2}. While vaccination efforts globally are underway, distribution efforts have been impeded in low- and middle-income countries. 

Additionally, the emergence of new viral variants like the Omicron variant has decreased the effectiveness of current vaccines in the prevention of COVID-19 spread \cite{3}.

The US economy has shown significant recovery so far; however, it has not recovered back to pre-pandemic levels, the certainty of improvement in this recovery is affected by these newly emerging COVID-19 variants \cite{8}. A recent surge in COVID, partially due to emerging variants, also affected job creation as fewer jobs than expected were added in December 2021 \cite{9}. Considering these socio-economic impacts of disease outbreaks like COVID-19, it is important to assess and predict socio-economic impacts and effects of mitigation and management approaches to help reduce outbreaks. Preliminary data and analysis supports positive U.S. socio-economic impact of a COVID-19 pre-screening solution utilizing Artificial Intelligence (AI). 

Sound has been used as one of the health indicators by clinicians and researchers, which often require skilled clinicians to interpret. Studies reveal that respiratory syndromes such as pneumonia, pulmonary diseases, and asthma \cite{10} \cite{11} \cite{12} can be diagnosed effectively using cough samples. Recently researchers have explored the possibility of using sound instead of imaging techniques like MRI, sonography, as they offer a cheap alternative. In \cite{13}, a crowd-sourced dataset comprising sounds from 7000 users was collected using both Android and web applications. Out of the 7000, only 200 users were tested positive for COVID-19. They aimed to use simple machine learning models like support vector machines (SVM) to discriminate the cough from users with COVID-19 from healthy users and users with asthma. In \cite{14}, a non-invasive, real-time framework adapted from the brain model by the MIT Center for Brain Minds and Machine designed for diagnosing Alzheimer's is used to pre-screen cough to identify COVID-19 by leveraging orthogonal audio biomarkers. In this study, 96 cough samples from users with bronchitis and 136 with pertussis are collected in addition to 2660 COVID positive and 2660 healthy cough samples.

Recurrent Neural networks are gaining popularity in speech recognition and audio analysis. Studies using the long short-term memory (LSTM) network models reveal that comparing cough and breath, voice is inefficient \cite{15}. 240 recordings from healthy individuals and 60 samples from COVID-19 patients collected from several hospitals in UAE are used in this study. Mouawad et al. exploited the non-linear phenomenon in voice and speech signals to capture the repetition pattern using time-series recurrence plots \cite{16}. They have used the data collected in partnership with Carnegie Mellon University and Voca.ai to train the machine learning models.

The DiCOVA challenge was initiated to accelerate researchers in COVID-19 diagnosis, and it uses respiratory acoustics \cite{17}. Two datasets derived from the crowd-sourced Coswara data \cite{18} are used in the DiCOVA challenge. The first dataset focuses only on cough sounds, and the second dataset contains a collection of audio recordings, including counting numbers, phonation of sustained vowels, and breath. AI models like logistic regression (LR), multi-layer perceptron (MLP), and random forest (RF) are analyzed on the cough, breath, vowel sounds, and speech samples. A group of researchers from the University of Cambridge combined the 11 most common symptoms with 384 features obtained from 16 frame-level descriptors using linear kernel SVM \cite{19}.

Pahar et al. from Stellenbosch University, South Africa, evaluated the popular machine learning approaches like SVM, k-nearest neighbor (kNN), LR, MLP, ResNet50, and LSTM, on the Coswara dataset and the SARS COVID-19 South Africa (Sarcos) dataset \cite{20} collected from all the six continents.

Erdoğan et al. \cite{21} used 595 positive and 592 negative COVID-19 cough samples from the Virufy dataset \cite{22}. They concluded that the performance of features based on deep learning is inferior, compared to traditional methods, due to limited data availability.

In \cite{23}, experiments are conducted to identify the most informative acoustic features as a baseline instead of the handcrafted features or Mel frequency Cepstral Coefficients. It is also demonstrated that robust features can be learned using the wavelet scattering transform amidst the noise in the data. 92.38\%, among the 1103 participants, were declared negative, with the remaining 7.62\% declared COVID-19 positive.

In \cite{24}, four classes of cough samples are used in this study, they are users tested COVID-19 (346), COVID-19 negative (346), healthy users with cough (101), and users without COVID but has pertussis cough (20). A collection of 813 samples from the University of Lleida, University of Cambridge, Virufy, and Pertussis datasets, are used. Compared to the latest machine learning methods, random forest performed better on the time-frequency features.
From this study, we can conclude that researchers worldwide are actively working towards a cost-effective AI/ML solution to pre-screen the infection from the cough samples collected from the user’s smartphone. However, the efficiency of these ML solutions is highly dependent on the balanced nature of the datasets and often in popular datasets healthy samples often outnumber the COVID-19 positive samples.

In this paper, we have utilized the costs associated with Hospital Acquired Infections (HAI) and the savings realized by preventing spread as a relevant comparable, as COVID-19 is also an Airborne Transmissible Disease. Our novel COVID-19 calculator measures the U.S. socio-economic impact of a COVID-19 AI pre-screening algorithm, and potential savings through early detection, facilitating the return to pre-COVID normalcy.  Calculation assumptions are based on a 1\% or 0.1\% degree of positive change. By using one’s own data and assumptions, and this model calculator, each user can customize the calculator based on values, weights, variables, and assumptions the user deems relevant for their local contexts or countries. 

The rest of the paper is organized as follows: we have discussed the various similar calculators developed for estimating the impact of COVID-19 in section 2, the proposed calculator is presented in section 3 and section 4 concludes the article. We have also summarized the sample calculations from other popular calculators in Appendix A.

\section{Discussion}
In considering this Article, a number of existing published articles, research, and calculators were reviewed:

\subsection{APIC - Association for Professionals in Infection Control \& Epidemiology (APIC) Algorithm \& Data Set \cite{4}}

APIC is the leading professional association for infection preventionists (IPs) with more than 15,000 members. APIC’s mission is to advance the science and practice of infection prevention and control.

The APIC Cost of Healthcare-Associated Infections Model is designed to demonstrate the costs associated with infections and the savings realized by preventing them. It also provides tables and graphs that describe the financial impact of infections at your healthcare institution. By using your own data, you can customize this report for your respective facility. If you don’t have your own data, APIC has provided data from national studies to estimate economic endpoints. Use of your healthcare organization’s data will reflect the financial impact of infections to your institution. 

\subsection{TMIT – APIC (Texas Medical Institute of Technology (TMIT) - Association for Professionals in Infection Control \& Epidemiology (APIC)) \cite{4}}

The TMIT-APIC Healthcare-Associated Infections Cost Calculator was developed in collaboration with the Texas Medical Institute of Technology (TMIT) and APIC. It provides an alternate method to determine the cost of healthcare-associated infections from the APIC cost calculator. The input to the TMIT-APIC calculator includes: hospital size category and the hospitals' infection data. The hospital size category is based on the number of beds, region and teaching status. And the infection data comprise of number of infections/year, excess costs/infection and excess length of stay (LOS)/infection. The calculator generates a report on infections such as surgical-site (SSI), ventilator-associated Pneumonia (VAP), catheter-associated urinary tract (UTI), central line-associated bloodstram (CLABSI), Methcillin-resistant Staphylococcus aureus (MRSA) and Clostridium difficile (C. diff.). 

\subsection{Education Week \cite{5}}

Education Week together with the Learning Policy Institute developed an interactive tool to calculate the percentage of revenue cuts for the 2019 - 2020 and 2020 - 2021 school year. They also accounted in the increased costs due the required high-speed internet for online education and increased number of days free or reduced price food are provided.

\subsection{St. Louis Fed Article \cite{6}}

The Bureau of Labor Statistics (BLS) uses surveys to gather data on the prices of goods and services purchased across the U.S., weights these prices by how much they contribute to the typical basket of expenditures, and then aggregates to form the consumer price index (CPI). Inflation is then measured as the rate of growth of the CPI over a specific period.
Alberto Cavallo, an economics professor at Harvard Business School, constructed expenditure weights for each month of 2020 using data from the Opportunity Insights Economic Tracker at Harvard University and Brown University as compiled by Raj Chetty and other economists \cite{cavallo}.

\begin{figure}[h!]
\begin{center}
\includegraphics[width=0.5\textwidth]{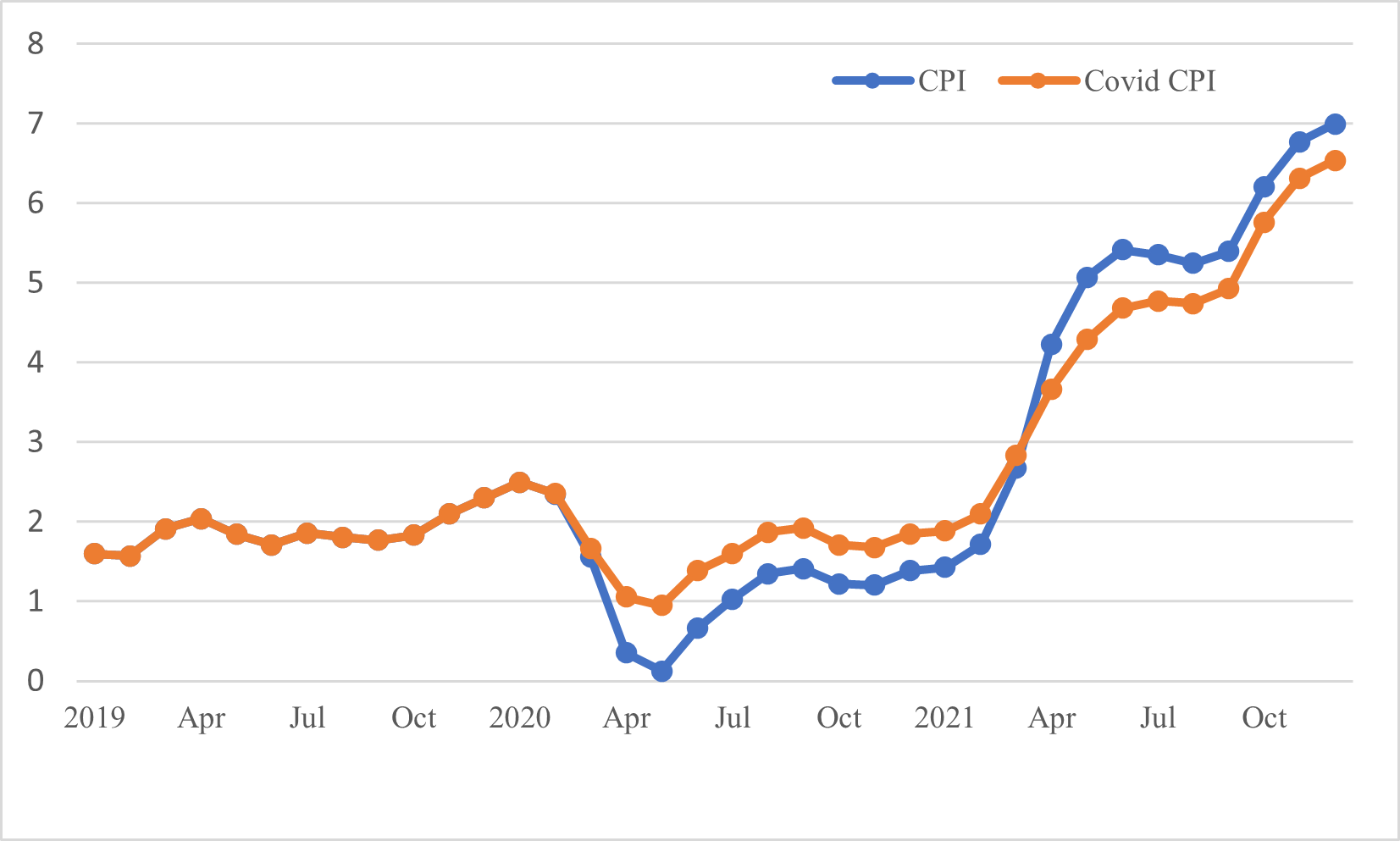}
\caption{The Pandemic’s Influence on the U.S. Inflation}
\label{inflation}
\end{center}
\end{figure}

Cavallo’s weights are, no doubt, imprecise but seem reasonable. For example, we know that consumer expenditures at restaurants fell sharply in 2020. It therefore seems likely that expenditures on Food at Home (Food Away from Home) likely increased (decreased) in 2020 and hence the true expenditure weights should be higher (lower) than normal for these categories. In fact, these weights are estimated to have increased (decreased) by 3.76 (3.07) percentage points during the month of April 2020, when stay-at-home orders were most severe for much of the country. Figure \ref{inflation} portrays the 12-month changes on the US COVID inflation for the years January 2019 to October 2021.

Having the estimated expenditure weights for 2020, we can then compare the official CPI-based inflation with an unofficial COVID CPI-based inflation that is easily accessible at Cavallo’s website.

\subsection{Medscape \cite{7}}

The COVID-19 Prognostic Tool estimates mortality rates in patients with COVID-19 and is adapted from Centers for Disease Control and Prevention (CDC) materials.

\section{Methods}
At present, we are not aware of any validated COVID-19 calculator that aggregates COVID value creation and have thus constructed this working model. The strength of our working model is that it leverages a number of existing models, and existing data points, and allows the user to customize the calculator based on values, weights, variables, and assumptions they deem relevant.  

Our novel model COVID-19 calculator measures U.S. socio-economic impact of COVID-19 AI/ML pre-screening algorithm, and potential savings through early detection, facilitating the return to pre-COVID normalcy.  Users may utilize criteria \& assumptions they are comfortable with.  For purposes of this illustration, assumptions are based on 1\% or 0.1\% degree of positive change. By using your own data and assumptions, and this model calculator, each user can customize the calculator based on values, weights, variables, and assumptions the user deems relevant.

The following factors were considered in designing the calculator:
\begin{itemize}
\setlength\itemsep{0.5pt}
    \item[-] Number of year cost(s) to bring the AI solution into the U.S. market.
    \item[-] Future potential through public/private partnership
    \item[-] Reduction in COVID-related healthcare expenses
    \item[-] Increase in GDP
    \item[-] Reduction due to COVID-related deaths
    \item[-] Jobs saved
    \item[-] Reduction in PCR testing Cost
    \item[-] Reduction in school-related expenses and delayed learning
    \item[-] Reduced COVID-related inflation
    \item[-] Any other measurable progress in returning back to pre-COVID normalcy
    \item[-] Any other additional criterion from the user
\end{itemize}

In table \ref{calculator}, we have summarized the steps involved for a sample calculation using the proposed calculator. The users can replace the sample calculations with actual data to determine the actual impact. We have provided the link to the calculator with sample calculations for better understanding.
\begin{table*}[h!]
\small
\centering
\caption{COVIDCal:  Potential U.S Value Creation Calculator considering a Variety of Indicators - Focused on the Public Good / ROI Methodology, utilizing a COVID-19 AI Pre-Screening Algorithm		\\		}  
\label{calculator}
\begin{tabular}{p{7.8cm}|c|c|p{2.2cm}}
\hline
Costs \& Value Creation	&	 Debits (a) - (d) &	 Credits (e) - (n) &	 Positive Value Creation (Credit - Debit)\\ \hline \hline

- (a) assume \# of Year cost(s) to bring the AI solution for the U.S. market [e.g., \$75,000,000.00] or other value from the user	&	 \$ 75,000,000.00 	& & \\ \hline	

- (b) additional future potential \$ through a public / private partnership for the U.S. market [e.g., \$ TBD] or other value from the user	&	 TBD 	& & \\ \hline	

 - (c) Additional criteria 1 from the user 	&	 TBD 	& & \\ \hline	
 - (d) Additional criteria 2 from the user 	&	 TBD 	& & \\ \hline	
 
 + (e) Reduction in COVID-related healthcare expenses for the U.S. market [e.g., assume a solution can contribute to a .1\% reduction at COVID-related healthcare expenses - \$50B loss per month, assume 12 months = \$600B, which means \$600M; assume conversion of loss = ten cents per dollar = \$60,000,000.00] or other value from the user	&	&	 \$ 60,000,000.00 & \\ \hline 	

+ (f) Increase in GDP for the U.S. market (e.g., assume a solution can contribute to an increase in GDP of the U.S. of 1/10th of 1\% of GDP [2021 forecasted US GDP \$21T. 0.04 attributed to COVID (\$840B). 1/10th of 1\% = \$840M] or other value from the user  & &			 \$ 840,000,000.00 	 & \\ \hline

+ (g) assume 1\% reduction of COVID related deaths in the U.S. [e.g., U.S. 790,000 deaths, assume 1\% reduction conversion is 7,930 lives per year = 23,790 lives / 3 years] or other value from the user 	& &		 TBD 	& \\ \hline
+ (h) assume .1\% savings of 22 million U.S. jobs lost = 22,000 jobs saved / year [e.g., U.S. = 66,000 jobs / 3 years] or other value from the user & &			 TBD 	& \\ \hline
+ (i) Reduction in U.S. PCR testing [users may add Antigen testing cost as well if they prefer] or other value from the user & &			 \$ 8,000,000.00 & \\ \hline	
+ (j) Reduction in U.S. school-related expenses \& delayed learning or other value from the user & &			 \$ 50,000,000.00 & \\ \hline	
+ (k) Reduction in COVID-related U.S. inflation [e.g., .1\% of 50 basis points] or other value from the user & & TBD 	& \\ \hline
+ (l) Measurable progress of U.S. returning to pre-COVID normalcy	& &		 TBD 	& \\ \hline
 + (m) Additional criteria 3 from the user & &			 TBD & \\ \hline	
 + (n) Additional criteria 4 from the user & &			 TBD & \\ \hline
 
Subtotal Debit (s)	&	 \$ 75,000,000.00 	& & \\ \hline	
Subtotal Credit (s)	&	&	 \$ 958,000,000.00 	 & \\ \hline
Positive Economic, Societal, and Health Benefits	& &	&		 \$ 883,000,000.00 \\ \hline \hline
\end{tabular}
\end{table*}

The COVIDCalc can be downloaded from: \url{https://github.com/virufy/covid\_calc}
\section{Conclusion}
The pace at which each country recovers from COVID shock varies from one another due to several factors such as increased domestic spending power, decreased consumption of contact-intensive services, labor shortages, etc. The surge due to variants from time to time adds up to other factors causing inflation.  Hence, it is important to evaluate or predict the socio-economic impact of disease outbreaks like COVID-19.
 
Several organizations and researchers have designed calculators to determine the cost incurred with the diseases and the associated savings by preventing them. Mostly, these calculators are industry-specific like the APIC cost calculator determines the impact on healthcare institutions and the Edweek calculator on the schools.
 
Recently, few AI/ML researchers have started exploring the usage of coughs collected from smartphones to pre-screen associated diseases like COVID-19, Asthma, Pertussis, etc. In this paper, we have designed a COVID-19 calculator that measures the socio-economic impact involved in the development of an AI/ML pre-screening solution and the associated savings through early detection. We have considered 1\% or 0.1\% degree of positive change in the calculations. We have enabled the users of this calculator to use their own data and assumptions to customize by assigning values, variables, weights, and assumptions the user deems fit.

\section*{Acknowledgements} 

We are very grateful to:

\begin{enumerate}
    \item Mark Parinas, CEO, CuraeSoft Corporation, for his thoughtful guidance on the socio-economic implications of our research, and his heartfelt commitment to the public good.
    \item The Covid Detection Foundation (dba Virufy) for their AI/ML efforts in support of our research.
    \item M. Pilanci is supported by the National Science Foundation (NSF), and the U.S. Army Research Office.
\end{enumerate}
 
\section*{Fair Use}
Subject to general limitations, some copyrighted material(s), such as the included figures, are deemed fair use(s) based on this paper’s intentions to stimulate creativity for the enrichment of the general public, comment, criticism, research, news reporting, teaching, and scholarship.  Cited work(s) are the property of the author(s) / copyright holders.

\appendix

\section{Sample Calculations from Other Popular Calculators}
\label{appendix}
\subsection{APIC Calculator}
The APIC calculator uses the following input to determine the potential savings: HAI incidence per 1000 patient days, mean attributable costs of HAIs, estimated cost to facility per 1000 patient days and potential decrease in HAI cases. 

\begin{table}[h!]
\small
\centering
\caption{Sample Calculation from the APIC calculator \cite{4} \\} 
\begin{tabular}{|p{4.5 cm}|c|}
\hline
Attribute                                        & Sample Value           \\ \hline \hline
HAI incidence per 1000 patient days              & 9.3                    \\ \hline
Mean attributable costs of HAIs                  & $ 13, 973 - $15,275    \\ \hline
Estimated cost to facility per 1000 patient days & $ 129,949 - $ 142, 057 \\ \hline
Potential decreases in HAI cases                 & upto 83\%              \\ \hline
Potential Savings                                & $ 107,858 - $117,907 \\ \hline \hline
\end{tabular}
\label{apicCal}

\end{table}
\subsection{TMIT - APIC Calculator}
\begin{figure*}[h!]
\begin{center}
\includegraphics[width=0.8\textwidth]{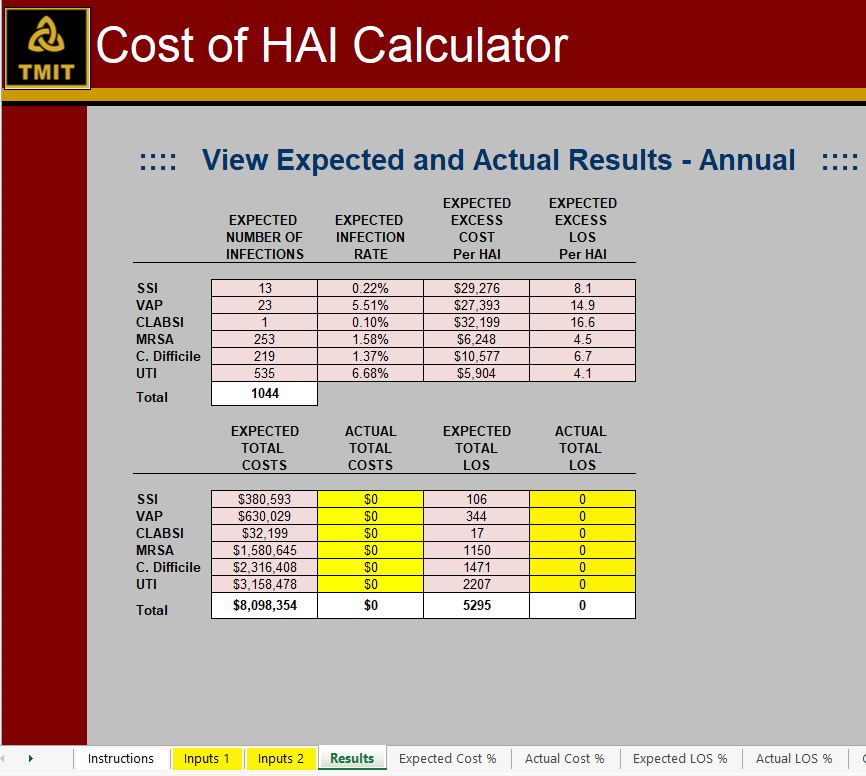}
\caption{Estimated HAI costs for a medium size hospital in the Northeast region with non-teaching status \cite{4}}
\label{haicost}
\end{center}
\end{figure*}

In fig. \ref{haicost}, the estimated cost for a medium size hospital in the Northeast region with non-teaching status is portrayed. The total annual medical and surgical admissions are assumed as 20,000 and 9000 respectively. The annual patient volume with ventilator, urinary catheter and central line are assumed as 1000, 8000 and 1000 respectively. Since the actual number of infections, excess cost per HAI and excess LOS per HAI are unknown, we have left it blank and are set as 0.

\subsection{Education Week's Calculator}
\begin{figure*}[h!]
\begin{center}
\includegraphics[width=0.8\textwidth]{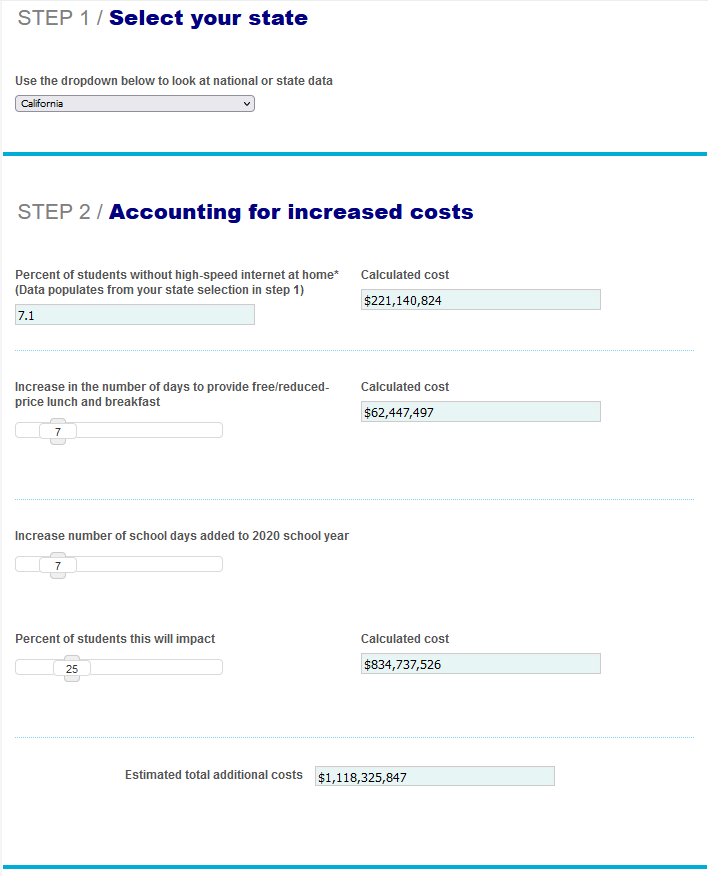}
\caption{Steps 1 - 2 of the Sample Calculation from the Education Week Calculator \cite{5}}
\label{edweek1}
\end{center}
\end{figure*}

\begin{figure*}[h!]
\begin{center}
\includegraphics[width=0.8\textwidth]{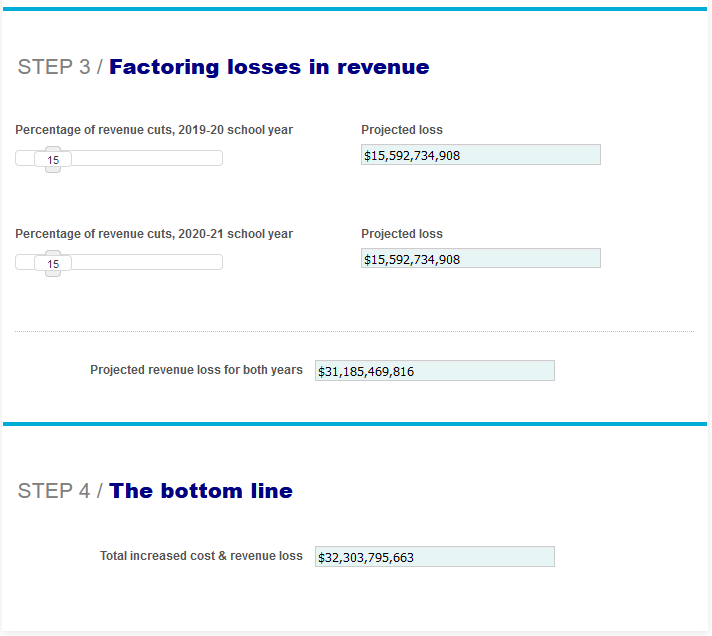}
\caption{Steps 3 - 4 of the Sample Calculation from the Education Week Calculator \cite{5}}
\label{edweek2}
\end{center}
\end{figure*}

Fig. \ref{edweek1} - \ref{edweek2}, show total increased cost and revenue loss calculated by the Education week calculator using the input like state, percent of students without high-speed internet at home, increase in the number of days to provide free or reduced price food, increase number of school days added to 2020 school year, percent of student that will be impacted, percentage of revenue cuts in 2019 - 2020 and 2020 - 2021 school year. 

\subsection{COVID-19 Prognostic Tool}
CDC's interim clinical guidance for managements of confirmed COVID-19 patients' data was used by this tool created by QxMD. It demonstrates that age is one of the strong risk factors in addition to pre-existing conditions like cancer, diabetes, hypertension, cardiovascular and respiratory diseases. However, researchers from the Johns Hopkins School of Medicine states that the case fatality rate is likely overestimated and so is the inflation based on mortality estimates \cite{25}.    
\begin{figure*}[h!]
\begin{center}
\includegraphics[width=0.8\textwidth]{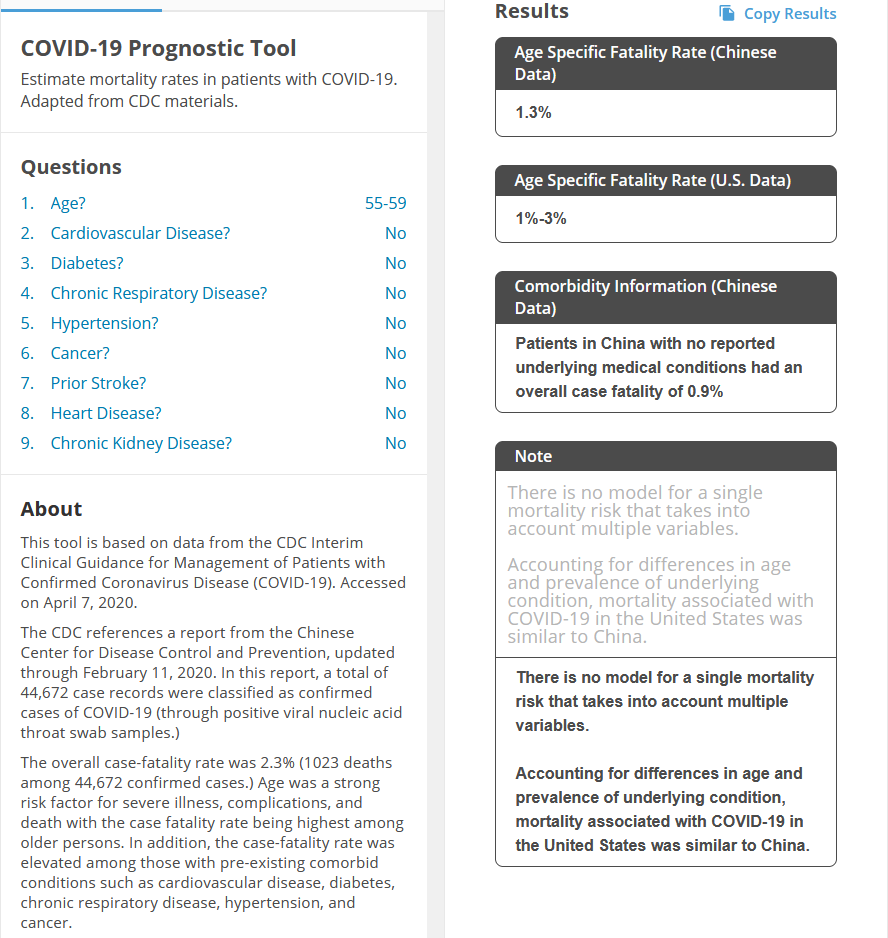}
\caption{COVID-19 Prognostic Tool \cite{7}}
\label{prognostic}
\end{center}
\end{figure*}

\end{document}